\documentclass[letterpaper, 10 pt, conference]{ieeeconf}  

\IEEEoverridecommandlockouts                              
\def\figurename{Fig. }

\usepackage[cmex10]{amsmath}
\usepackage{eufrak}
\usepackage{multirow}
\usepackage[caption=false,font=footnotesize]{subfig}
\usepackage[pdftex]{graphicx}
\DeclareGraphicsExtensions{.pdf,.jpeg,.png,.jpg}
\usepackage{cite}

\title{\LARGE \bf
Memory Management for Real-Time Appearance-Based \\ Loop Closure Detection
}

\author{Mathieu Labb\'e and Fran{\c c}ois Michaud \\
Department of Electrical and Computer Engineering\\
Universit\'e de Sherbrooke, Sherbrooke, QC, CA J1K 2R1\\
\{mathieu.m.labbe, francois.michaud\}@usherbrooke.ca}

\begin{document}

\maketitle
\thispagestyle{empty}
\pagestyle{empty}

\begin{abstract}
Loop closure detection is the process involved when trying to find a match between the current and a previously visited locations in SLAM. 
Over time, the amount of time required to process new observations increases with the size of the internal map, which may influence real-time processing.
In this paper, we present a novel real-time loop closure detection approach for large-scale and long-term SLAM. Our approach is based on a memory management method that keeps computation time for each new observation under a fixed limit. Results demonstrate the approach's adaptability and scalability using four standard data sets. 
\end{abstract}

\section{INTRODUCTION}
Autonomous robots operating in real life settings must be able to navigate in large, unstructured, dynamic and unknown spaces. Simultaneous localization and mapping (SLAM) \cite{thrun05} is the capability required by robots to build and update a map of its operating environment and to localize itself in it. A key feature in SLAM is to recognize previously visited locations. This process is also known as loop closure detection, referring to the fact that coming back to a previously visited location makes it possible to associate this location with another one recently visited.

For global loop closure detection approaches \cite{Newman06, Cummins08a,Angeli08b, botterill2011bag, konolige2010view}, a general way to detect a loop closure is to compare a new location with the ones associated with previously visited locations. If no match is found, then a new location is added. A problem with this approach is that the amount of time required to process new observations increases with the number of locations in the map. If this becomes larger than the acquisition time, a delay is introduced, leading to an obsolete map. In addition, a robot operating in large areas for a long period of time will ultimately build a very large map, making updating and processing the map difficult to achieve in real-time. 

Our solution is to manage the locations used to detect loop closures based on their occurrences. The idea is simple (inspired from works in psychology \cite{atkinson1968human, baddeley1997human}): it consists of keeping the most recent and frequently observed locations in the robot's Working Memory (WM), and transferring the others into a Long-Term Memory (LTM). When a match is found between the current location and one stored in WM, associated locations stored in LTM can be remembered and updated.

This paper describes our memory management approach to accomplish appearance-based loop closure detection with real-time constraint for long-term SLAM. Processing time is the criterion used to limit the number of locations kept in WM, and therefore satisfy real-time constraint. Our memory management mechanism ensures satisfaction of real-time constraint independently of the scale of the mapped environment. For this reason, we named our approach Real-Time Appearance-Based Mapping (RTAB-Map). Trials conducted with four standard data sets demonstrate real-time capabilities of our approach under various conditions.   

The paper is organized as follows. Section \ref{sec:related_work} summarizes related work on loop closure detection for large-scale SLAM. Section \ref{sec:system_description} describes our approach. Section \ref{sec:results} presents experimental results and Section \ref{sec:discussion} shows a qualitative comparison of our approach with similar ones in the community. Section \ref{sec:conclusion} concludes the paper.

\section{RELATED WORK}
\label{sec:related_work}
For most of the probabilistic SLAM approaches \cite{thrun05, Folkesson07, Blanco08, callmer2008tree, Paz08, Pinies08, schleicher2010real}, loop closure detection is done locally, i.e., matches are found between new observations and a limited region of the map, so that process can be accomplished in real-time and at 30 Hz \cite{davison2007monoslam}.
The size of the region is determined by the uncertainty associated with the robot's position. Such approaches fail if the estimated position is erroneous.
Newman et al. \cite{Newman06} indicate that in real world situations, it is very likely that events will cause gross errors in the estimated position.
Therefore, it is preferable to do loop closure detection without using an estimated position.

Global loop closure detection consists of comparing a new observation to observations stored in the map, and to add this observation only if no match is found.
Vision is the sense generally used because of the distinctiveness of the features extracted from the environment \cite{Milford08, Newman05, Tapus08}, although successful large-scale mapping using features extracted using laser range finder data is possible \cite{Bosse07}. 
For vision-based mapping, the bag-of-words \cite{sivic2003video} approach is commonly used \cite{Cummins08a, Angeli08c, botterill2011bag, konolige2010view} and has shown to perform loop closure detection in real-time for paths of up to 1000 km \cite{cummins2009highly}.
The bag-of-words approach consists of representing each image by visual words taken from a dictionary.
The visual words are usually made from local feature descriptors, such as Speeded-Up Robust Features (SURF) \cite{Bay08}.
Each word keeps a link to images it is associated to, making image retrieval efficient over large data set. 
However, as a robot explores new areas over long periods of time, the number of images to compare increases linearly, making it eventually impossible to conduct the matching process in real-time. 
The number of comparisons can be decreased by considering only a selection of previously acquired images (referred to as key images) for the matching process, while keeping detection performance nearly the same to using all images \cite{booij2009efficient}. However, complexity still increases linearly with the number of key images.
This phenomenon can also occur in close and dynamic environments in which the robot frequently visits the same locations: perceptual aliasing, changes that can occur in dynamic environments or the lack of discriminative information may affect the ability to recognize previously visited locations, leading to the addition of new locations in the map and consequently affecting the satisfaction of real-time constraints \cite{glover10}.
To limit the growth of the number of images to match, pruning \cite{milford2010persistent} or clustering \cite{konolige2009towards} methods can be used to consolidate portions of the map which exceed a spatial density threshold. 
This limits growth over time, but not according to the size of the explored environment.

\section{REAL-TIME APPEARANCE-BASED MAPPING (RTAB-MAP)}
\label{sec:system_description}
The objective of our work is to provide a solution independent of time and size, to achieve real-time loop closure detection for long-term operation in large environments. The idea resides in only using a certain number of locations for loop closure detection so that real-time constraint can be satisfied, while still having access to the locations of the entire map when necessary. When the number of locations in the map makes processing cycle time for finding matches greater than a real-time threshold, our approach transfers locations less likely to cause loop closure detection from the robot's WM to LTM, so that they do not take part in the detection of loop closures. However, if a loop closure is detected, neighbor locations can be retrieved and brought back into WM to be considered in loop closure detections. Note that our approach can be seen as a dynamic way to generate and to dynamically maintain a constant number of key locations in WM, which differs from other key frame approaches \cite{booij2009efficient} where the number of key locations would increase over time and space.

Transferred locations in LTM are not used to find a match with the current one. Therefore, some loops may not be detected, making it important to carefully choose which locations to transfer to LTM. A naive approach is to use a first-in first-out (FIFO) policy, pruning the oldest locations from the map. However, this sets a limit on the maximum number of locations memorized in sequence when exploring an environment: if the real-time limit is reached before loop closure can be detected, pruning the older locations will make it impossible to eventually find a match. Another approach would be to randomly pick locations to transfer, but it is preferable to keep in WM the locations that are more susceptible to be revisited. Therefore, our proposed approach evaluates the number of times a location has been matched or consecutively viewed (referred to as Rehearsal) to set its weight, making it possible to identify the locations seen more frequently than others and that are more likely to cause loop closure detections. When transfer occurs, the location with the lowest weight is selected. If there are locations with the same weight, then the oldest one is selected to be transferred. 

Similar to \cite{Angeli08b}, locations in the Short-Term Memory (STM) are not used for loop closure detection, to avoid loop closure detection on locations that have just been visited (the last location always looks similar to the most recent ones). STM size is fixed based on the robot velocity and the rate at which the locations are acquired. When the number of locations reaches STM size limit, the oldest location in STM is moved into WM.

\begin{figure}[!t] 
\centering 
\includegraphics[width= 2.6in]{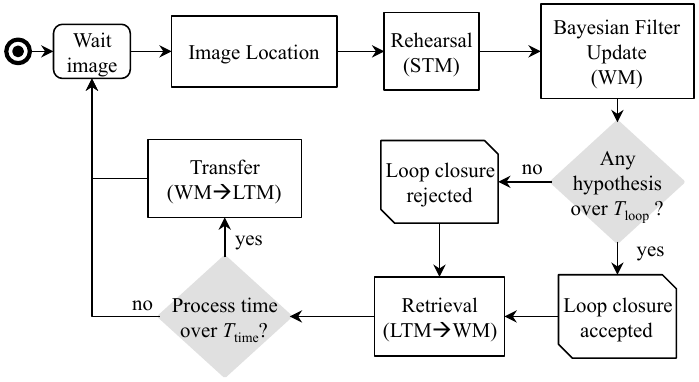} 
\caption[Flow chart of the loop closure detection approach]{Flow chart of our memory management loop closure detection processing cycle.} 
\label{fig_detector_process} 
\end{figure}

\figurename \ref{fig_detector_process} illustrates the overall loop closure detection process which is explained in details in the following sections.

\subsection{Image Location}
\label{sec:location}

The bag-of-words approach \cite{sivic2003video} is used to create a signature $z_t$ of an image acquired at time $t$.  
An image signature is represented by a set of visual words contained in a visual dictionary incrementally constructed online using a randomized forest of kd-trees \cite{muja_flann_2009}. 
Speeded-Up Robust Features (SURF) \cite{Bay08} are extracted from the image as the visual words. 
A location $L_t$ is then created with signature $z_t$, a weight initialized to 0 and a bidirectional link in the graph with $L_{t-1}$. Note that the dictionary contains only words of locations in WM and STM.

\subsection{Rehearsal}
\label{sec_memory_rehearsal}

To update the weight of the acquired location, $L_t$ is compared to the ones in STM from the most recent ones to the latest, and similarity $\mathfrak{s}$ is evaluated using (\ref{eq:similarity}) :

\begin{equation}	
	\label{eq:similarity}
	\mathfrak{s}(z_t, z_c)=\left\{\begin{matrix}
N_{\mathrm{pair}}/N_{z_t}, & \textrm{if } N_{z_t}\geq N_{z_c} \\ 
N_{\mathrm{pair}}/N_{z_c}, & \textrm{if } N_{z_t} < N_{z_c}
\end{matrix}\right. 
\end{equation}

\noindent where $N_{\mathrm{pair}}$ is the number of matched word pairs between the compared location signatures, and $N_{z_t}$ and $N_{z_c}$ are the total number of words of signature $z_t$ and the compared signature $z_c$ respectively. 
If $\mathfrak{s}(z_t, z_c)$ is higher than a fixed similarity threshold $T_{\mathrm{rehearsal}}$, $L_c$ is merged into $L_t$ and Rehearsal stops. In Rehearsal, only words of $z_c$ are kept in the merged signature : $z_t$ is cleared and $z_{c}$ is copied into $z_t$.
To complete the merging process, the weight of $L_t$ is increased by the one of $L_c$ plus one, the neighbor links of $L_c$ are added to $L_t$ and $L_c$ is deleted from STM.

\subsection{Bayesian Filter Update}
\label{sec:bayes}

The role of the discrete Bayesian filter is to keep track of loop closure hypotheses by estimating the probability that the current location $L_t$ matches one of an already visited location stored in the WM. 
Let $S_t$ be a random variable representing the states of all loop closure hypotheses at time $t$. 
$S_t = i$ is the probability that $L_t$ closes a loop with a past location $L_i$, thus detecting that $L_t$ and $L_i$ represent the same location. $S_t = -1$ is the probability that $L_t$ is a new location.
The filter estimates the full posterior probability $\boldsymbol {p}(S_t|L^t)$ for all $i=-1,...,t_{n}$, where $t_{n}$ is the time index associated with the newest location in the WM, expressed as follows \cite{Angeli08c}: 

\begin{equation}	
	\label{eq:bayes}
\boldsymbol{p}(S_t|L^t) = \eta 
\underbrace{
\boldsymbol{p}(L_t|S_t) 
}_\textrm{Observation}
\underbrace{
\displaystyle\sum\limits_{i=-1}^{t_{n}} 
\underbrace{
\boldsymbol{p}(S_t|S_{t-1}=i) 
}_\textrm{Transition}
p(S_{t-1}=i|L^{t-1})
}_\textrm{Belief}
\end{equation}

\noindent where $\eta$ is a normalization term and $L^{t} = L_{-1},...,L_{t}$. Note that the sequence of locations $L^t$ includes only the locations contained in WM and STM, thus $L^t$ changes over time as new locations are created and when some locations are retrieved from LTM or transferred to LTM, deviating from the classical Bayesian filtering where such sequences are fixed.

The observation model $\boldsymbol{p}(L_t|S_t)$ is evaluated using a likelihood function $\mathfrak{L}(S_t|L_t)$ : the current location $L_t$ is compared using (\ref{eq:similarity}) to locations corresponding to each loop closure state $S_t=j$ where $j=0,..,t_n$, giving a score $s_j=\mathfrak{s}(z_t,z_j)$. Each score is then normalized using the difference between the score $s_j$ and the standard deviation $\sigma$, normalized by the mean $\mu$ of all non-null scores, as in (\ref{eq:normalizedscore}) \cite{Angeli08c}  :
\begin{equation}
	\label{eq:normalizedscore}
p(L_t|S_t=j) = \mathfrak{L}(S_t=j|L_t)= \left\{\begin{matrix}
\frac{s_j-\sigma}{\mu}, & \textrm{if } s_j\geq \mu+\sigma \\ 
1, & \textrm{otherwise.}
\end{matrix}\right. 
\end{equation}
\noindent For the new location probability $S_t=-1$, the likelihood is evaluated using (\ref{eq:virtualPlaceLikelihood}) :
\begin{equation}	
	\label{eq:virtualPlaceLikelihood}
	p(L_t|S_t=-1) = \mathfrak{L}(S_t=-1|L_t) = \frac{\mu}{\sigma} + 1
\end{equation}
\noindent where the score is relative to $\mu$ on $\sigma$ ratio. If $\mathfrak{L}(S_t=-1|L_t)$ is high, meaning that $L_t$ is not similar to a particular one in WM (when $\sigma<\mu$), then $L_t$ is more likely to be a new location.

The transition model $\boldsymbol {p}(S_t|S_{t-1}=i)$ is used to predict the distribution of $S_t$, given each state of the distribution $S_{t-1}$ in accordance with the robot's motion between $t$ and $t-1$. Combined with $p(S_{t-1}=i|L^{t-1})$ (i.e., the recursive part of the filter) this constitutes the belief of the next loop closure. 
The transition model is expressed as in \cite{Angeli08c}.
\begin{enumerate}
  \item $p(S_t=-1|S_{t-1}=-1)=0.9$, the probability of a new location event at time $t$ given that no loop closure occurred at time $t-1$.
  \item $p(S_t=i|S_{t-1}=-1)=0.1/N_{\mathrm{WM}}$ with $i\in[0;t_n]$, the probability of a loop closure event at time $t$ given that no loop closure occurred at $t-1$. $N_{\mathrm{WM}}$ is the number of locations in WM of the current iteration.
  \item $p(S_t=-1|S_{t-1}=j)=0.1$ with $j\in[0;t_n]$, the probability of a new location event at time $t$ given that a loop closure occurred at time $t-1$ with $j$.
  \item $p(S_t=i|S_{t-1}=j)$, with $i,j\in[0;t_n]$, the probability of a loop closure event at time $t$ given that a loop closure occured at time $t-1$ on a neighbor location. The probability is defined as a discretized Gaussian curve centered on $j$ and where values are non-null for exactly eight neighbors (for $i=j-4,...,j+4$) and their sum are 0.9.
\end{enumerate}

\subsection{Loop Closure Hypothesis Selection}
\label{sec:loop_selection}
When $\boldsymbol {p}(S_t|L^t)$ has been updated and normalized, the highest hypothesis of $\boldsymbol {p}(S_t|L^t)$ greater than the loop closure threshold $T_{\mathrm{loop}}$ is selected. Because a loop closure hypothesis is generally diffused to its neighbors (neighbor locations share some similar words between them), for each probability $p(S_t=i|L_t)$ where $i\geq0$, the probabilities of its neighbors in the range defined by the transition model $p(S_t=i|S_{t-1}=j)$ are summed together. 
Note that for a selection to occur, a minimum of $T_{\mathrm{minHyp}}$ locations must be in WM: when the graph is initialized, the first locations added to WM have automatically a high loop closure probability (after normalization by (\ref{eq:bayes})), therefore wrong loop closures could be accepted if the hypotheses are over $T_{\mathrm{loop}}$. 

If there is a loop closure hypothesis $p(S_t=i|L^t)$ where $i\geq0$ is over $T_{\mathrm{loop}}$, 
the hypothesis $S_t=i$ is then accepted, $L_t$ is merged with the old location $L_i$: 
the weight of $L_t$ is increased by the one of $L_i$ plus one and the neighbor links of $L_i$ are added to $L_t$. After updating $L_t$, unlike Rehearsal, $L_i$ is not immediately deleted from the memory. For the constancy of the Bayesian filter with the next acquired images, the associated loop closure hypothesis $S_i$ must be evaluated until $L_i$ is not anymore in the neighborhood of the highest loop closure hypothesis, after which it is deleted.

\subsection{Retrieval}
\label{sec:retrieval}

Neighbors of location $L_i$ in LTM are transferred back into WM if $p(S_t=i|L^t)$ is the highest probability. 
Because this step is time consuming, a maximum of two locations are retrieved at each iteration (chosen inside the neighboring range defined in Section \ref{sec:bayes}). 
The visual dictionary is updated with the words associated with the signatures of the retrieved locations. 
Common words from the retrieved signatures still exist in the dictionary, so a simple reference is added between these words and the corresponding signatures. 
For words that are not present anymore (because they were removed from the dictionary when the locations were transferred), they are matched to those in the dictionary to find if more recent words represent the same SURF features. 
This step is particularly important because the new words added from the new signature $z_t$ may be identical to the previously transferred words. 
For matched words, the old words are replaced by the new ones in the retrieved signatures. All the references in the LTM are changed and the old words are permanently removed.
If some words are still unmatched, they are simply added to the dictionary.

\subsection{Transfer}
\label{sec:transfer}

When processing time becomes greater than $T_{\mathrm{time}}$, the oldest location with the lowest weight is transferred from WM to LTM. 
$T_{\mathrm{time}}$ must be set to allow the robot to process the perceived images in real-time. 
Higher $T_{\mathrm{time}}$ means that more locations (and implicitly more words) can be kept in the WM, and more loop closure hypotheses can be kept to better represent the overall environment.
$T_{\mathrm{time}}$ must therefore be set according to the robot's CPU capabilities, computational load and operating environment. 
If $T_{\mathrm{time}}$ is set to be higher than the image acquisition time, the algorithm intrinsically uses an image rate corresponding to $T_{\mathrm{time}}$, with 100\% CPU usage.
As a rule of thumb, $T_{\mathrm{time}}$ can be set to about 200 to 400 ms smaller to the image acquisition rate at 1 Hz, to ensure that all images are processed under the image acquisition rate, even if the processing time goes over $T_{\mathrm{time}}$ ($T_{\mathrm{time}}$ then corresponds to the average processing time of an image by our algorithm).
So, for an image acquisition rate of 1 Hz, $T_{\mathrm{time}}$ could be set between 600 ms to 800 ms.

Because the most expensive step of our algorithm is to rebuild the visual dictionary, process time can be regulated by changing the dictionary size, which indirectly influences the WM size.
A signature of a location transferred to LTM removes its word references from the visual dictionary. 
If a word does not have reference to a signature anymore, it is transferred into LTM. 
While the number of words transferred from the dictionary is less than the number of words added from the new or retrieved locations, more locations are transferred. 
At the end of this process, the dictionary size is smaller than before the new words from the new and retrieved locations were added, thus reducing the time required to update the dictionary for the next acquired image. 
Saving the transferred locations into the database is done asynchronously using a background thread, leading to a minimal time overhead.
Note also that to be able to evaluate appropriately loop closure hypotheses using the discrete Bayesian filter, the retrieved locations of the current iteration are not allowed to be transferred.

\section{RESULTS}
\label{sec:results}
\label{results}

Performance of our approach is evaluated in terms of precision-recall metrics \cite{cummins2009highly}. Precision is the ratio of true positive loop closure detections to total detections. Recall is defined as the ratio of true positive loop closure detections to the number of ground truth loop closures. 

Because the precision is very important in loop closure detection, all results shown in this paper were done at 100\% precision, thus no false positive was accepted.

Using a MacBook Pro 2.66 GHz Intel Core i7, we conducted tests with the following community data sets: NewCollege and CityCentre data sets \cite{Cummins08a}, and Lip6Indoor and Lip6Outdoor data sets \cite{Angeli08c}. 
The NewCollege and CityCentre data sets contain images acquired from two cameras (left and right). Because our approach only takes one image as input in its current implementation, the two images were merged into one, resulting in one $1280 \times 480$ image. 
With our bounded data sets, we could use the LTM database directly in computer's RAM, but we preferred to use it on the hard drive to simulate a more realistic setup for timing performances.

Experimentation of our loop closure detection approach is done using parameters $T_{\mathrm{rehearsal}}$ set to 20\%, STM size to 25, $T_{\mathrm{loop}}$ to 10\% and $T_{\mathrm{minHyp}}$ to 15.
The same set of parameters is used over the different data sets to evaluate adaptability of our approach without optimization for specific conditions. 
For this reason, we did not do an exhaustive search for the best parameters; instead, they were set empirically to avoid false positives (precision of 100\%) over all data sets while giving good overall recall performances.

For each data set, three experiments were done with different $T_{\mathrm{time}}$ to show the effect of memory management on the recall performances. When $T_{\mathrm{time}}=\infty$, all locations are kept in WM, and therefore loop closure detection is done over all previously visited locations, thus giving optimal results. These results are used to compare those when real-time constraint is applied, i.e., with $T_{\mathrm{time}}$ set based on the guideline presented in Section \ref{sec:transfer} ($T_{\mathrm{time}}=1.4 \, \mathrm{s}$ for images acquired at 0.5 Hz, i.e., every 2 seconds; or $T_{\mathrm{time}}=0.7 \, \mathrm{s}$ for images acquired at 1 Hz), and with $T_{\mathrm{time}}$ set smaller to such guideline to show the influence of a smaller WM on recall and timing performances. Smaller $T_{\mathrm{time}}$ could be used but with lower recall performances if some locations with a high weight are transferred before the robot returns in the area of those locations. Remember also that WM must have a minimum of $T_{\mathrm{minHyp}}$ locations to efficiently evaluate loop closure probabilities.

\begin{table*}[!t]
\renewcommand{\arraystretch}{1.3}
\caption{Experimental conditions and results of RTAB-Map for the community data sets}
\label{table_community_results}
\centering
\begin{tabular}{l|ccc|ccc|ccc|ccc}
\hline
Data set & 
\multicolumn{3}{c|}{NewCollege} & 
\multicolumn{3}{c|}{CityCentre} & 
\multicolumn {3}{c|}{Lip6Indoor} &
\multicolumn {3}{c}{Lip6Outdoor} \\
\hline
\# images & 
\multicolumn{3}{c|}{1073} & 
\multicolumn{3}{c|}{1237} & 
\multicolumn {3}{c|}{388} & 
\multicolumn {3}{c}{531} \\
Image rate & 
\multicolumn{3}{c|}{$\approx$0.5 Hz} & 
\multicolumn{3}{c|}{$\approx$0.5 Hz} & 
\multicolumn {3}{c|}{1 Hz} & 
\multicolumn {3}{c}{0.5 Hz} \\
Image size & 
\multicolumn{3}{c|}{$1280\times480$} & 
\multicolumn{3}{c|}{$1280\times480$} & 
\multicolumn {3}{c|}{$240\times192$} & 
\multicolumn {3}{c}{$240\times192$} \\
\hline
$T_{\mathrm{time}}$ (s) & 
$\infty$ & \textbf{1.4} & 0.7 & 
$\infty$ & \textbf{1.4} & 0.7 &
$\infty$ & \textbf{0.7} & 0.1 & 
$\infty$ & \textbf{1.4} & 0.4 \\
\hline
Max WM size (locations) & 
524 & \textbf{299} & 147 &
724 & \textbf{366} & 159 &
115 & \textbf{115} & 80 & 
280 & \textbf{280} & 102\\
Max dictionary size  ($\times10^3$ words) & 
169 & \textbf{99} & 52 &
224 & \textbf{106} & 52 &
12 & \textbf{12} & 8 &
89 & \textbf{89} & 31 \\
Max iteration time (s) & 
2.95 & \textbf{1.75} & 0.95 & 
3.57 & \textbf{1.83} & 1.00 &
0.185 & \textbf{0.185} & 0.13 &
1.34 & \textbf{1.34} &  0.51 \\
Precision (\%) & 
100 & \textbf{100} & 100 &
100 & \textbf{100} & 100 &
100 & \textbf{100} & 100 &
100 & \textbf{100} & 100 \\
Recall (\%) & 
85 & \textbf{84} & 85 &
80 & \textbf{79} & 79 &
86 & \textbf{86} & 79 &
77 & \textbf{77} & 81\\
\hline
\end{tabular}
\end{table*}

\begin{figure}[!t] 
\centering 
\includegraphics[width= 2.2in]{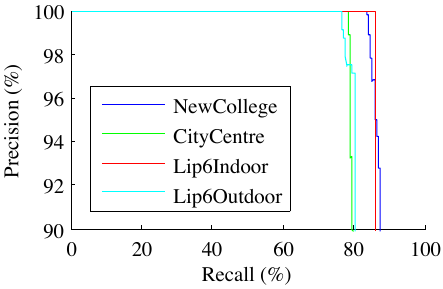} 
\caption[Precision-recall curves]{Precision-recall curves for each data set.} 
\label{fig:Precision_Recall} 
\end{figure}

Table \ref{table_community_results} summarizes experimental conditions and results. Results (shown in bold) are the ones done with $T_{\mathrm{time}}$ set to values suggested accordingly to image rate. For the NewCollege and CityCentre data sets, RTAB-Map achieves similar performances to when no locations are transferred to LTM ($T_{\mathrm{time}}=\infty$), but respecting real-time constraints. For the Lip6Indoor and Lip6Outdoor, because processing time never reached the real-time constraint, results at $T_{\mathrm{time}}=\infty$ are the same than with the experiments with $T_{\mathrm{time}}=0.7 \, \mathrm{s}$ and $T_{\mathrm{time}}=1.4 \, \mathrm{s}$ respectively. Using a smaller $T_{\mathrm{time}}$ does not influence much recall performances, athough in some cases (like with the Lip6Outdoor data set) better results are obtained because some noisy locations, which could incorrectly have high likelihoods, are not in WM, resulting in higher hypotheses on correct previously visited locations.  
Typical precision-recall curves are shown in \figurename \ref{fig:Precision_Recall}, where higher recall performances at 100\% precision are the better.
Compared to other approaches, RTAB-Map achieves better recall performances at 100\% precision than those reported using FAB-MAP in \cite{Cummins08a} (47\% for NewCollege, 37\% for CityCentre) and those in \cite{Angeli08c} (80\% for Lip6Indoor, 71\% for Lip6Outdoor).

\begin{figure}[!t] 
\centering 
\includegraphics[width= 2.6in]{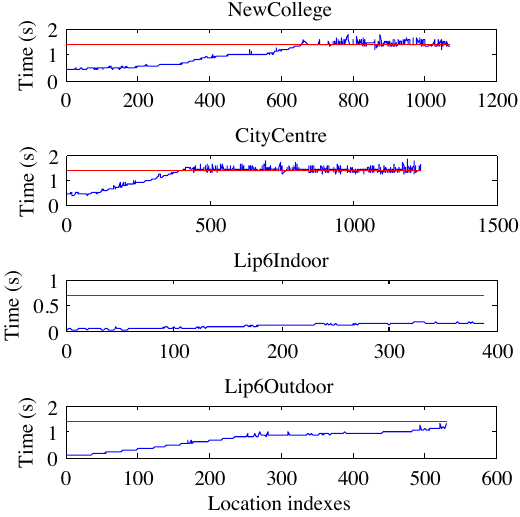} 
\caption[Processing time]{Processing time in terms of locations processed over time for each data set. The time limit $T_{\mathrm{time}}$ used is shown as a red line.} 
\label{fig:time_results_community} 
\end{figure}

\figurename \ref{fig:time_results_community} shows processing time results for each data set. For the NewCollege and CityCentre data sets, the processing time for each new acquired image increases until $T_{\mathrm{time}}$ is reached, then it remains around $T_{\mathrm{time}}$ and below the real-time limit (set to 0.5 Hz, i.e., 2 seconds). For the Lip6Indoor and Lip6Outdoor data sets, the time taken never reached $T_{\mathrm{time}}$: the smaller slope after processing half of the Lip6Outdoor data set is caused by many loop closures detected, and then memory size increased very slowly.

\section{DISCUSSION}
\label{sec:discussion}

\begin{table*}[!t]
\renewcommand{\arraystretch}{1.3}
\caption{Comparison with loop closure detection approaches.}
\label{table_comparison}
\centering
\begin{tabular}{ccccccc}
\hline
Approach & Signature type & Adaptability & Usability & Repeatability & Hypothesis quality & Real-time satisfaction \\
\hline
IAB-Map \cite{Angeli08c} & SIFT+H histograms & + & - & + & - & limited \\
IAB-Map \cite{Angeli08b} & SIFT & + & + & + & - & limited \\
FAB-Map \cite{Cummins08a} & SURF & - & - & + & +& limited \\
Acc. FAB-MAP \cite{Cummins08b} & SURF & - & - & + & + & limited \\
FAB-MAP 2.0 \cite{cummins2009highly} & SURF & - & - & + & + & limited \\
RTAB-Map & SURF & + & + & - & - & unlimited \\
\hline
\end{tabular}
\end{table*}

Table \ref{table_comparison} compares the proposed loop closure detection approach (RTAB-Map) with Incremental Appearance-Based Mapping (IAB-Map) \cite{Angeli08c, Angeli08b} and FAB-MAP based approaches \cite{Cummins08a, Cummins08b, cummins2009highly}, in terms of:

adaptability (the ability of the method to adapt itself to the environment), 
usability (ease of use in SLAM), 
repeatability (determinism), 
hypothesis quality (robustness to perceptual aliasing) and
real-time satisfaction (under a maximum of data acquired or not).

RTAB-Map and IAB-Map approaches always have a visual words dictionary which represents well the environment, updating itself while the environment changes. FAB-MAP uses a fixed dictionary, pre-trained with images representing the target environment, required because of its more complex method of hypothesis filtering. The same dictionary could be used over different environments but performances are not optimal \cite{Cummins08a}, limiting adaptability. A pre-computed dictionary is however useful to provide priors on the probability of appearance of a word (using its frequency in the dictionary), giving better likelihood values.
While a pre-computed dictionary would give better likelihood values (for its target environment), we preferred to use a incremental one because it is simpler to setup, increasing usability.
FAB-MAP has also lower usability because it requires non-overlapping images derived from a pre-filtering (using odometry) done over the acquired images. The two other methods manage overlapping images automatically, discarding (like IAB-Map) or using (like RTAB-Map) them, thus making them easier to use. 
IAB-Map and FAB-MAP are deterministic, i.e., they lead to the same recall performance when processing the same data set. Because RTAB-Map transfers locations based on computation time, which is affected by other processes running on the robot's computer, a small latency while accessing the hard drive (which plays the role of the database) may cause changes in WM, leading to small differences in performances. This is not a limitation however, because in real life settings the environment is dynamic, causing a similar effect.
Robustness to perceptual aliasing is handled in all approaches, but FAB-MAP handles it at the word level rather than at the signature level.
Finally, regarding real-time processing for long-term operation, our approach is the only one that is not limited by a maximum of images acquired, a very important criterion in our opinion to achieve long-term SLAM.
All approaches are limited by the memory available on the robot, although RTAB-Map is limited by the hard drive size and not the computer's RAM size.

\section{CONCLUSION}
\label{sec:conclusion}
Results presented in this paper suggest that RTAB-Map, a loop closure detection approach based on a memory management mechanism, is able to meet real-time constraints needed for long-term online mapping. 
While keeping a relatively constant number of locations in WM, real-time processing is achieved for each new image acquired. 
Retrieval is a key feature that allows our approach to reach adequate recall ratio even when transferring a high proportion of the perceived locations. 
In future work, we want to test the approach over a larger data set to evaluate more properly the impact when some locations can be seen much more often than others, a situation which could affect the recall ratio over time.

\section{ACKNOWLEDGMENT}
F. Michaud holds the Canada Research Chair in Mobile Robotics and Autonomous Intelligent Systems. Support for this work is provided by the Natural Sciences and Engineering Research Council of Canada, the Canada Research Chair program and the Canadian Foundation for Innovation.

\bibliographystyle{IEEEtran}
\bibliography{IEEEabrv,../../../Papers/References}

\end{document}